# A Convolutional Cost-Sensitive Crack Localization Algorithm for Automated and Reliable RC Bridge Inspection

S.O. Sajedi & X. Liang
*Department of Civil, Structural and Environmental Engineering, University at Buffalo, New York, United States.*

ABSTRACT: Bridges are an essential part of the transportation infrastructure and need to be monitored periodically. Visual inspections by dedicated teams have been one of the primary tools in structural health monitoring (SHM) of bridge structures. However, such conventional methods have certain shortcomings. Manual inspections may be challenging in harsh environments and are commonly biased in nature. In the last decade, camera-equipped unmanned aerial vehicles (UAVs) have been widely used for visual inspections; however, the task of automatically extracting useful information from raw images is still challenging. In this paper, a deep learning semantic segmentation framework is proposed to automatically localize surface cracks. Due to the high imbalance of crack and background classes in images, different strategies are investigated to improve performance and reliability. The trained models are tested on real-world crack images showing impressive robustness in terms of the metrics defined by the concepts of precision and recall. These techniques can be used in SHM of bridges to extract useful information from the unprocessed images taken from UAVs.

## 1 INTRODUCTION

Reduced time to recovery is one of the fundamental characteristics of resilient systems. Bridge infrastructures, as critical components in a transportation system, play an important role in large communities. Such structures should maintain their functionality in harsh environments especially after extreme events (e.g., earthquakes). Given the number of aging infrastructure across the united states, structural health monitoring (SHM) techniques have been widely used to periodically monitor the condition of bridges. Visual inspections are one of the most common ways of condition assessment where this task is conventionally performed by dedicated teams. There are several drawbacks for human inspections. Having dedicated teams for this purpose requires time and monetary resources that may not be readily available after disasters. Moreover, bridges are commonly built in harsh geographical locations to facilitate transportation. Critical structural components may not be easily accessible for manual investigations of damage. That being said, most visual inspections are inaccurate and biased (Phares 2001) while reliable information about bridge condition is essential to the decision makers. To address these issues, automated SHM has been the topic of interest in many studies (Spencer et al. 2019). Camera-equipped unmanned aerial vehicles can be effectively used in this regard. However, obtaining useful information from raw images is still challenging.

With the rapid progress of the research in the field of artificial intelligence, recent deep learning models have been capable of classifying object within raw images. Proposed algorithms are mainly designed to detect common objects such as pedestrians, roads, etc. With a similar approach, damage detection can be performed in SHM to extract meaningful information from raw images. Recently there has been an interest in implementing deep learning for the given task (e.g., Liang 2018, Yang et al. 2018). Detecting cracks, as one of the dominant damage types in reinforced concrete (RC) bridges is an important task that can be effectively handled with deep learning. However, the performance of data-driven image classifiers is highly dependent on the training data. In pictures taken from the cracked structural components, unlike most object detection problems, there exist a significant imbalance between background and crack pixels. This imbalance may result in models that perform well in classifying background pixel while showing poor performance in identifying cracks. Nonetheless, the correct prediction of classes that correspond to damage (e.g., cracks) is an equally or more important objective in SHM.

In this paper, a pixel-wise crack segmentation algorithm is proposed to label individual pixels in real-world images. The deep learning model is inspired by SegNet (Badrinarayanan et al. 2015), a successful, yet computationally efficient architecture. In the following sections, the effect of different hyperparameters for the task of crack segmentation is studied. Moreover, utilizing Bayesian optimization, three different strategies are investigated to improve the model's robustness against severe class imbalance.

## 2 CRACK SEGMENTATION MODEL

In this paper, model training and evaluation are performed on a series of real-life unprocessed images. *Crack Forest* (Shi et al. 2016) is a publicly available dataset that includes 118 images with ground truth labeling of cracks and background pixels. This dataset is randomly shuffled with 80%-20% splits, respectively, for training and testing of the model. Moreover, 20% of the training data is held out for the validation which is utilized in evaluating the objective function of Bayesian optimization. The task of crack segmentation is performed using a fully convolutional encoder-decoder neural network. The deep learning architecture is inspired by SegNet which accepts colored 320×480 input images (as in *Crack Forest*) and outputs softmax probabilities of crack and background classes for individual pixels. These probabilities will be later modified or directly used in the decision rule.

Considering the significant imbalance in the distribution of existing classes, global mean accuracy may not be a proper metric to monitor the segmentation performance. For example, high global accuracies may result from a model that labels all pixels as background while the main focus of this study (and SHM) is to provide an accurate prediction of crack patterns. To avoid misleading evaluations, the concepts of precision ($P$) and recall ($R$) are adopted with the following definitions:

$$P = \frac{t^p}{t^p + f^p} \tag{1}$$

$$R = \frac{t^p}{t^p + f^n} \tag{2}$$

where $t^p$ is the true positive as the number of crack pixels identified correctly, $f^p$ is the total number of background pixels that are misclassified as crack, and $f^n$ is the number of crack pixels misclassified as background. To better distinguish the difference between one from the other, one may consider the information presented in Table 1.

Table 1. Different types of pixel segmentation outcomes

| Type of pixel classification | Ground truth label | Predicted label |
|---|---|---|
| $t^p$ | Crack | Crack |
| $f^p$ | Background | Crack |
| $f^n$ | Crack | Background |
| $t^n$ | Background | Background |

The metric *P* shows that among all pixels classified as crack, how many actually belong to this label. In contrast, the percentage of crack pixels in the ground truth that is correctly detected compared with the ones that were missed is evaluated with the recall (*R*). It can be implied that recall is a measure of completeness in crack segmentation. Theoretically speaking, a high precision value does not necessarily correspond to a high recall ratio while an ideal model will present, i.e., $P = R =1$. To provide an extreme example, labeling all pixels as crack will yield to $R =1$ where the precision is poor. In this case, although $f^n = 0$, all the background pixels are misclassified which is an excessive overestimation. Given that the majority of pixels have the ground truth label of background, $f^p$ will be a relatively large number in (1) which yields to a very low accuracy (i.e., precision) for this example. In common datasets, there is usually a trade-off between the two metrics. $F$1 score can also be used as a single scalar that contains combined information about both precision and recall and can have values between 0 and 1 as two extremes. This metric is obtained by calculating the harmonic mean of $P$ and $R$ as follows:

$$F1 = \frac{2PR}{P+R} \tag{3}$$

It should be noted that the output of a neural network is the prediction probability of each label (e.g. crack and background). To label each pixel, one may select the one with the highest probability. However, in binary classification, it is possible to define certain threshold for the cracks. For example, instead of taking the label with max probability, one may consider all the pixels that have at least 20% probability of being crack as such. In this case, $P$ and $R$ can be calculated considering different probability thresholds to segment cracks. Given these values, different combinations of ($R$, $P$) can be obtained by considering a series of thresholds. This will result in a precision-recall curve. The area under the curve is known as mean precision accuracy (*MPA*) which can also be used as a single metric to evaluate the crack segmentation performance. $P$, $R$, $F$1-score, and *MPA* are later used in this paper to investigate the effects of hyperparameters in crack segmentation.

As mentioned earlier, training the deep learning architecture with uniform weights (UW) for all observations and then taking the maximum a-posteriori probabilities (MAPs) can result in a model with near perfect global accuracy. However, precision and recall values may be low for the classification of crack pixels. In this paper, UW-MAP will be utilized as the baseline strategy while two other approaches are investigated to improve the crack segmentation performance. Training observations can be weighed depending on the true class of each pixel. For example, median frequency weight (MFW) assignment proposed by Eigen & Fergus (2015) has been originally adopted in SegNet. It should be noted that such weights differ from the network learnable parameters and should be selected as hyperparameters prior to the training. With a different approach in a recent publication, Chan et al. (2019) modified the decision rule by using the maximum likelihood (ML). In this case, softmax probabilities are modified based on the prior distribution of damage classes in each pixel. Therefore, maximum likelihood probabilities are used to identify a pixel as crack or background. Depending on the location, the frequency of each class is illustrated in Figure 1. By pixel-specific normalization, prior probabilities used in the ML approach are obtained for crack and background classes. In this paper, we investigate these two strategies and their effect by referring to them as MFW-MAP and UW-ML, respectively. The first term implies the way that the training observations are weighted while the second one refers to the adopted decision rule.

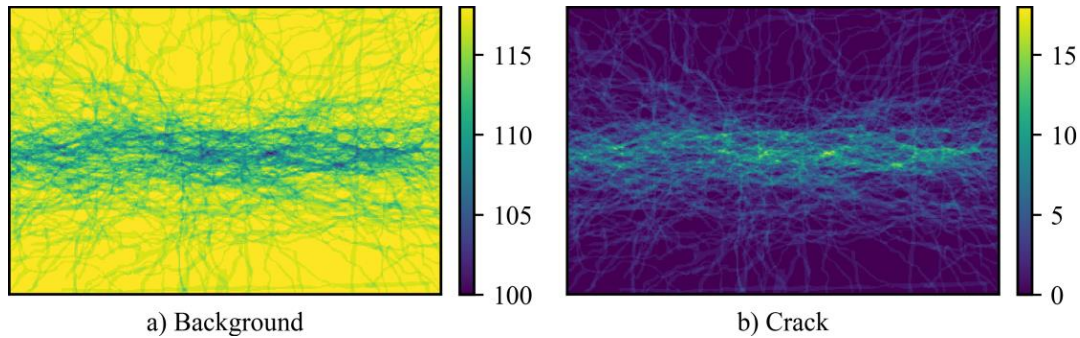

Figure 1. The frequency of two classes in the Crack Forest dataset

## 3 TUNING HYPERPARAMTERS

Learnable parameters of the deep learning model are the sliding kernel weights that are set through training in different epochs. Model training is accelerated by running epochs on an NVIDIA GeForce GTX 1080 GPU @ 8 GHz with 2560 CUDA cores, using Keras API (Chollet 2018). 80 epochs are found to be appropriate for training with the mini-batch size of 4 images. This deep learning architecture includes 4 pairs of encoder-decoder computation blocks. Padding, strides, the number of extracted filters, etc. are selected in a similar fashion to the original implementation of SegNet. However, given the input-output shapes of the crack segmentation model, proper adjustments are made. The architecture of the segmentation deep learning model is illustrated in Figure 2.

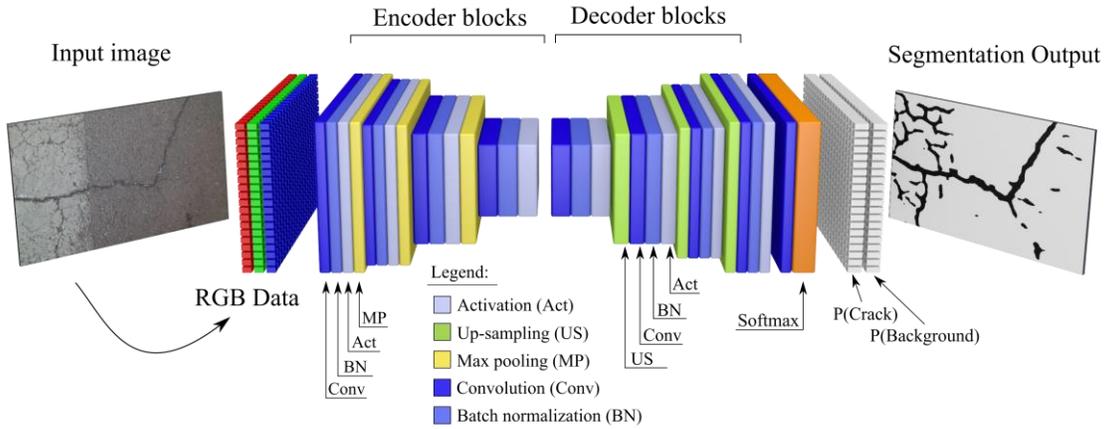

Figure 2. The Deep learning used in the semantic segmentation of cracks

Considering the default hyperparameters for each optimizer, model is independently trained for UW and MFW strategies. After completing the training process, the best set of learnable parameters were selected for further investigations. The criterion to select such parameters is the minimum validation obtained from the cross-entropy loss function. To minimize the loss, stochastic gradient descent (*SGD*), *RMSprop*, *Adagrad*, *Adadelta*, *Adam*, *Adamax* and *Nadam* optimizers are investigated. Assuming the default hyperparameters in the Keras API, a case study is performed to compare the performance of models trained with different optimizers. Given the three imbalance techniques mentioned earlier, average values of the performance metrics are documented for each weight optimizer. Results are shown in Figures 3 and 4. It should be noted that the *P*, *R*, *F*1-score, and *MPA* are calculated by assuming the crack as the positive class.

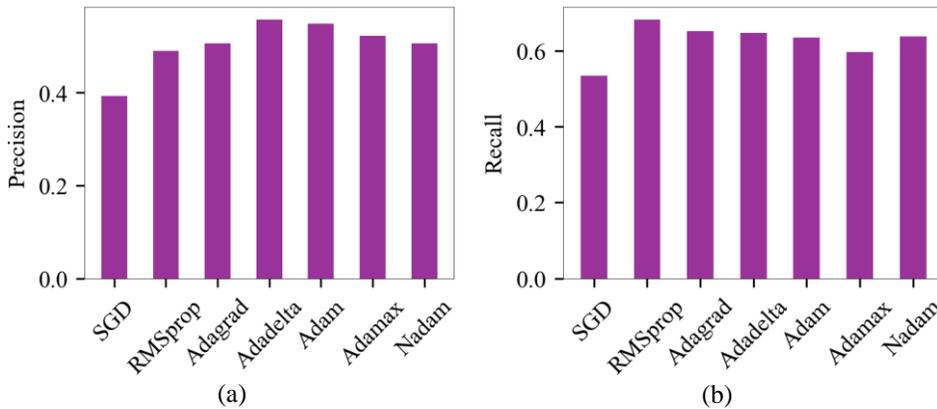

Figure 3. Precision and recall values for different Solvers (average value of UW-MAP, UW-ML, and MFW-MAP)

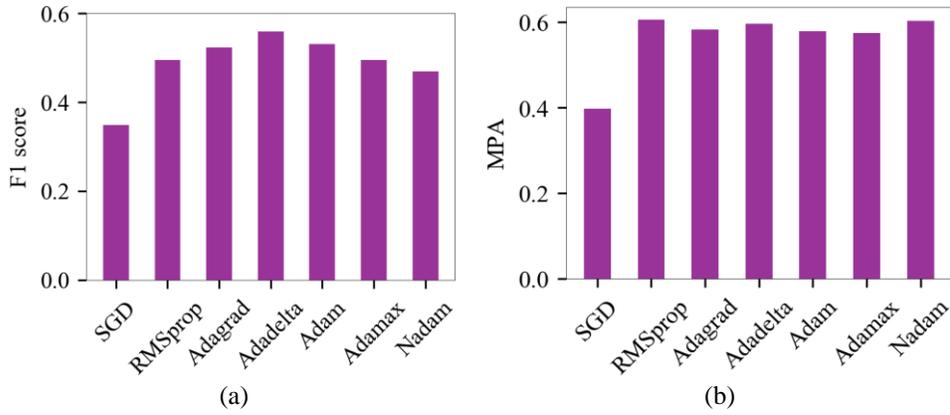

Figure 4. Performance of different optimizers (average value of UW-MAP, UW-ML, and MFW-MAP)

It can be observed that *Adadelta* yields relatively better performance compared with the others. However, the default learning hyperparameters of this algorithm may be tuned for enhanced performance. Bayesian optimization (Snoek et al. 2012) is used to adjust *Adadelta*'s hyperparameters to maximize *MPA* as the objective function. In this case, optimization is performed individually for UW-MAP, UW-ML and MFW-MAP strategies. The earlier metrics are documented for both crack and background pixels where the results are shown in Figure 5. Moreover, the precision-recall curves used to calculate *MPA* are shown in Figure 6.

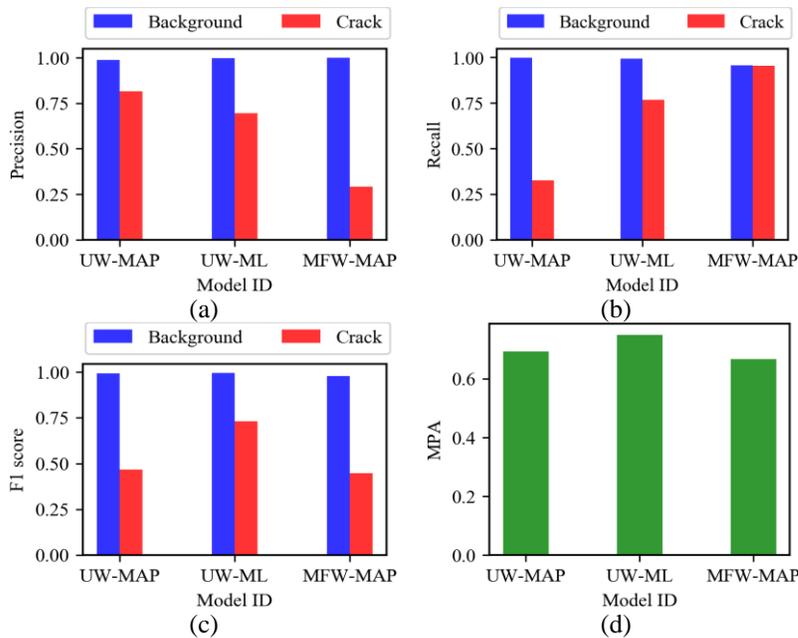

Figure 5. Comparison of the crack segmentation strategies after Bayesian optimization

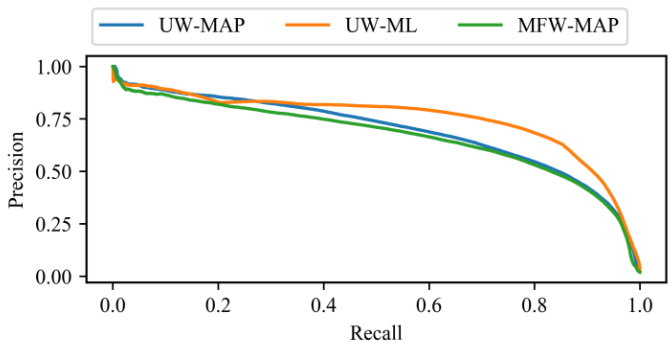

Figure 6. The precision recall curve for three different strategies

For the background pixels, as expected, the performance of the deep learning models is nearly perfect with respect to both precision and recall. From the baseline strategy, recall is relatively small, indicating that the majority of pixels corresponding to crack are not detected; however, it has the highest precision. In contrast, MFW-MAP shows a conservative prediction of cracks where recall values are the highest but with the lowest precision of the three strategies. UW-ML shows the best *F1*-score and *MPA* with moderate precision and recall values. The test set examples of detecting cracks in real-world images are provided in Figure 7.

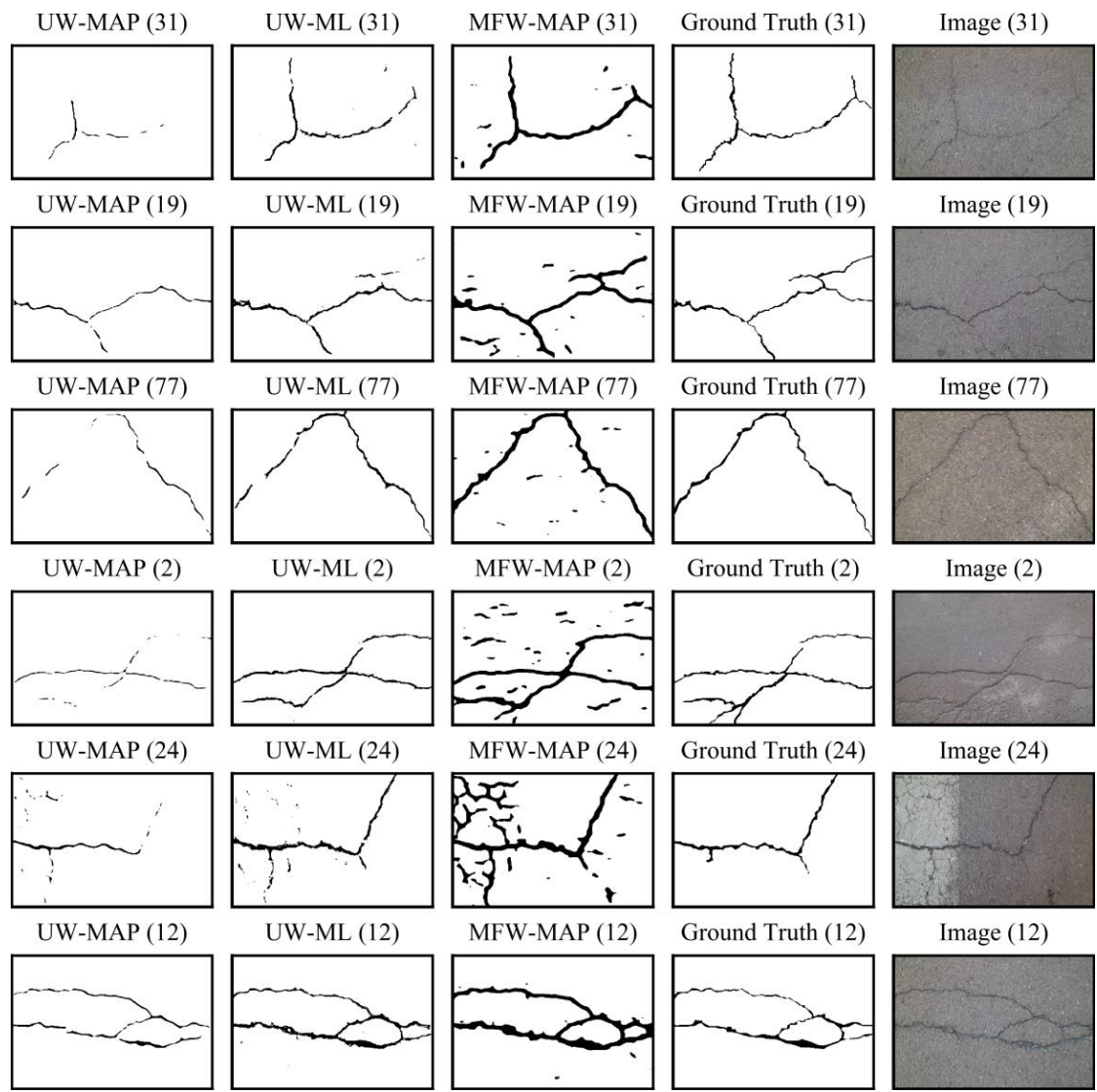

Figure 7. Crack segmentation results for example test observations

The first three columns in this figure represent the segmentation results by the three strategies. Ground truth segmentation from the original *Crack Forest* dataset is also provided. The numbers inside the parenthesis indicate to the image ID in *Crack Forest*. An interesting finding of this study is that in some cases (e.g., image 24), there exist cracks that are missed by the human-generated ground truth. However, the deep learning models, more or less, are able to detect such patterns. Some of these cracks are located on a more complex background (white paint). Yet, the performance is reasonable. The predictions from MFW-MAP are more conservative compared to the other methods while the predicted crack widths commonly appear thicker than ground-truth. In addition, misclassified background pixels (e.g., stains) are more likely to occur in this approach. In contrast, the predictions from the UW-MAP appear relatively incomplete as inferred from the recall values shown in Figure 5.b. Overall, the baseline model and MFW-MAP show inferior performance for, respectively, recall and precision metrics. Modifying the decision rule (UW-ML) provides better performance with a reasonable trade-off in precision and recall.

## 4 CONCLUSIONS

Reliable information is critical in maintaining the structural health of bridges. Automated SHM techniques are developed to obtain such information in an efficient manner. Surface cracks are a useful source of information regarding certain structural defects. However, processing raw real-world images may be challenging and biased when human inspections are performed. This paper proposes a fully automated semantic segmentation model to effectively detect cracks and overcome the challenge of the highly imbalanced dataset. Three strategies of UW-MAP, UW-ML, and MFW-MAP are considered for the task of crack segmentation. Moreover, Bayesian optimization is utilized to tune the hyperparameters for enhanced robustness. It is shown that by changing the training weights and modifying the decision rule, segmentation can be performed more effectively. Such improvements are investigated using *P*, *R*, *F*1-score, and *MPA*. UW-ML strategy shows better results compared with the others while maintaining a reasonable performance considering these metrics. Given the robustness of the deep learning models for the unseen raw images, these techniques can be effectively used for automated SHM inspection of bridge infrastructure.